\documentclass[aoasstyle, noinfoline]{imsart}
\usepackage[letterpaper, margin=1.35in]{geometry}
\usepackage{graphicx}	

\usepackage{natbib}
\usepackage{url}            
\usepackage{booktabs}       
\usepackage{todonotes}
\usepackage{amsmath, amssymb, bm}

\usepackage{graphicx}	
\usepackage{caption}
\usepackage{subcaption}
\usepackage{comment}
\usepackage{siunitx}
\usepackage{multirow}
\usepackage{booktabs}

\usepackage[nice]{nicefrac}

\usepackage{algorithm}
\usepackage[noend]{algpseudocode}

\usepackage{tikz}
\usetikzlibrary{fit,positioning,calc}
\definecolor{light}{RGB}{220, 188, 188}
\definecolor{mid}{RGB}{185, 124, 124}
\definecolor{dark}{RGB}{143, 39, 39}
\definecolor{highlight}{RGB}{0, 255, 0}
\definecolor{gray10}{gray}{0.1}
\definecolor{gray20}{gray}{0.2}
\definecolor{gray30}{gray}{0.3}
\definecolor{gray40}{gray}{0.4}
\definecolor{gray60}{gray}{0.6}
\definecolor{gray70}{gray}{0.7}
\definecolor{gray80}{gray}{0.8}
\definecolor{gray90}{gray}{0.9}
\definecolor{gray95}{gray}{0.95}
\definecolor{comment}{gray}{0.50}

\DeclareMathOperator*{\argmin}{arg\,min}

\newcommand{\bg}{\boldsymbol{g}}
\newcommand{\bz}{\boldsymbol{z}}
\newcommand{\blambda}{{\boldsymbol{\lambda}}}
\newcommand{\bC}{\boldsymbol{C}}
\newcommand{\bJ}{\boldsymbol{J}}
\newcommand{\bH}{\boldsymbol{H}}

\newcommand{\bs}{\boldsymbol{s}}

\newcommand{\bmm}{\boldsymbol{m}}
\newcommand{\bff}{\boldsymbol{f}}
\usepackage[parfill]{parskip}

\definecolor{light}{RGB}{0, 100, 100}
\usepackage[colorlinks,linkcolor=light,citecolor=light,urlcolor=light]{hyperref}

\newlength{\parvspace}
\begin{document}

\begin{frontmatter}
\title{Reducing Reparameterization Gradient Variance}
\runtitle{Reducing Reparameterization Gradient Variance}

\begin{aug}
  \author{\fnms{Andrew C.}  \snm{Miller}\corref{}\thanksref{t1}\ead[label=e1]{acm@seas.harvard.edu}\ead[label=u1,url]{http://andymiller.github.io}},
  \author{\fnms{Nicholas J.} \snm{Foti}\thanksref{t2}\ead[label=e2]{nfoti@uw.edu}},
  \author{\fnms{Alexander} \snm{D'Amour}\thanksref{t4}\ead[label=e4]{alexdamour@berkeley.edu}},
  \and
  \author{\fnms{Ryan P.}  \snm{Adams}\thanksref{t3}\ead[label=e3]{rpa@seas.harvard.edu}}

  \thankstext{t1}{Harvard University, \printead{e1}, \printead{u1}} 
  \thankstext{t2}{University of Washington, \printead{e2}}
  \thankstext{t3}{Harvard University, Google Brain \printead{e3}}
  \thankstext{t4}{UC Berkeley \printead{e4}}
  \runauthor{A.C.~Miller et al.}
\end{aug}

\begin{abstract}
Optimization with noisy gradients has become ubiquitous in statistics and machine learning.
Reparameterization gradients, or gradient estimates computed via the ``reparameterization trick,'' represent a class of noisy gradients often used in Monte Carlo variational inference (MCVI).
However, when these gradient estimators are too noisy, the optimization procedure can be slow or fail to converge.
One way to reduce noise is to use more samples for the gradient estimate, but this can be computationally expensive.
Instead, we view the noisy gradient as a random variable, and form an inexpensive approximation of the generating procedure for the gradient sample.
This approximation has high correlation with the noisy gradient by construction, making it a useful control variate for variance reduction.
We demonstrate our approach on non-conjugate multi-level hierarchical models and a Bayesian neural net where we observed gradient variance reductions of multiple orders of magnitude (20-2{,}000$\times$).

\end{abstract}

\end{frontmatter}


\section{Introduction}
\label{sec:intro}

Representing massive datasets with flexible probabilistic models has been central to the success of many statistics and machine learning applications, but the computational burden of fitting these models is a major hurdle.
For optimization-based fitting methods, a central approach to this problem has been replacing expensive evaluations of the gradient of the objective function with cheap, unbiased, stochastic estimates of the gradient.
For example, stochastic gradient descent using small mini-batches of (conditionally) i.i.d.\ data to estimate the gradient at each iteration is a popular approach with massive data sets.
Alternatively, some learning methods sample directly from a generative model or approximating distribution to estimate the gradients of interest, for example, in learning algorithms for implicit models~\cite{mohamed2016learning, tran2017deep} and generative adversarial networks~\cite{arjovsky2017wasserstein,goodfellow2014generative}.

Approximate Bayesian inference using variational techniques (variational inference, or VI) has also motivated the development of new stochastic gradient estimators, as the variational approach reframes the integration problem of inference as an optimization problem~\cite{blei2016variational}.
VI approaches seek out the distribution from a well-understood variational family of distributions that best approximates an intractable posterior distribution.
The VI objective function itself is often intractable, but recent work has shown that it can be optimized with stochastic gradient methods that use Monte Carlo estimates of the gradient \cite{kingma2013auto,ranganath2014black,rezende2014stochastic}, 
We call this approach Monte Carlo variational inference (MCVI).
In MCVI, generating samples from an approximate posterior distribution is the source of gradient stochasticity.
Alternatively, \emph{stochastic variational inference} (SVI)~\cite{hoffman2013stochastic} and other stochastic optimization procedures induce stochasticity through data subsampling; MCVI can be augmented with data subsampling to accelerate computation for large data sets.

The two commonly used MCVI gradient estimators are the \textit{score function gradient}~\cite{ranganath2014black} and the \textit{reparameterization gradient}~\cite{kingma2013auto,rezende2014stochastic}.  Broadly speaking, score function estimates can be applied to both discrete and continuous variables, but often have high variance and thus are frequently used in conjunction with variance reduction techniques.  On the other hand, the reparameterization gradient often has lower variance, but is restricted to continuous random variables. See \citet{ruiz2016generalized} for a unifying perspective on these two estimators.
Like other stochastic gradient methods, the success of MCVI depends on controlling the variance of the stochastic gradient estimator.

In this work, we present a novel approach to controlling the variance of the \emph{reparameterization gradient estimator} in MCVI.
Existing MCVI methods control this variance na\"ively by averaging several gradient estimates, which becomes expensive for large data sets and complex models,  with error that only diminishes as~$O(1/\sqrt{N})$.
Our approach exploits the fact that, in MCVI, the randomness in the gradient estimator is completely determined by a known Monte Carlo generating process;
this allows us to leverage knowledge about this generating process to de-noise the gradient estimator.
In particular, we construct a cheaply computed control variate based on an analytical linear approximation to the gradient estimator.
Taking a linear combination of a na\"ive gradient estimate with this control variate yields a new estimator for the gradient that remains unbiased but has lower variance.
We apply the idea to Gaussian approximating families and measure the reduction in gradient variance under various conditions. 
We observe a 20-2{,}000 $\times$ reduction in variance of the gradient norm in some conditions, and much faster convergence and more stable behavior of optimization traces.

\section{Background}
\label{sec:background}

\vspace{-.5em}
\paragraph{Variational Inference} 
Given a model,~${p(\bz, \mathcal{D}) = p(\mathcal{D} | \bz) p(\bz)}$, of data~$\mathcal{D}$ and parameters/latent variables~$\bz$, the goal of VI is to approximate the posterior distribution~$p(\bz | \mathcal{D})$. 
VI approximates this intractable posterior distribution with one from a simpler family,~${\mathcal{Q} = \{q(\bz; \blambda), \blambda \in \boldsymbol{\Lambda} \}}$, parameterized by \emph{variational parameters}~$\blambda$.  
VI procedures seek out the member of that family,~${q(\cdot; \blambda) \in \mathcal{Q}}$, that minimizes some divergence between the approximation~$q$ and the true posterior~$p(\bz | \mathcal{D})$.

Variational inference can be framed as an optimization problem, usually in terms of Kullback-Leibler (KL) divergence, of the following form
\begin{align*}
	\blambda^* &= \argmin_{\blambda \in \Lambda} \text{KL}(q(\bz; \blambda)~||~p(\bz | \mathcal{D})) 
	= \argmin_{\blambda \in \Lambda} \mathbb{E}_{\bz \sim q_\blambda}\left[ \ln q(\bz; \blambda) - \ln p(\bz | \mathcal{D}) \right]\,. 
\end{align*}
The task is to find a setting of $\blambda$ that makes $q(\bz ;\blambda)$ close to the posterior $p(\bz | \mathcal{D})$ in KL divergence.\footnote{We use $q(\bz; \blambda)$ and $q_\blambda(\bz)$ interchangeably.}
Directly computing the KL divergence requires evaluating the posterior itself; therefore, VI procedures use the \emph{evidence lower bound} (ELBO) as the optimization objective
\begin{align}
	\mathcal{L}(\blambda) &= \mathbb{E}_{\bz \sim q_\blambda}\left[ \ln p(\bz, \mathcal{D}) - \ln q(\bz; \blambda) \right] 
\end{align}
which, when maximized, minimizes the KL divergence between $q(\bz;\blambda)$ and $p(\bz|\mathcal{D})$. 

To maximize the ELBO with gradient methods, we need to compute the gradient of the ELBO,~$\bg_\blambda$.
The gradient inherits the ELBO's form as an expectation, which is in general an intractable quantity to compute.
In this work, we focus on \emph{reparameterization gradient estimators} (RGEs) computed using the \emph{reparameterization trick}.
The reparameterization trick exploits the structure of the \emph{variational data generating procedure} --- the mechanism by which~$\bz$ is simulated from~$q_\blambda(\bz)$. 
To compute the RGE, we first express the sampling procedure from $q_\blambda(\bz)$ as a differentiable map
\begin{align}
	\epsilon &\sim q_0(\epsilon) && \text{ independent of $\blambda$ } \\
	\bz      &= \mathcal{T}(\epsilon; \blambda) && \text{ differentiable map }
\end{align}
where the initial distribution $q_0$ and $\mathcal{T}$ are jointly defined such that~$\bz \sim q(\bz;\blambda)$ has the desired distribution. 
As a simple concrete example, if we set $q(\bz; \blambda)$ to be a diagonal Gaussian $\mathcal{N}(\bmm_\blambda, \text{diag}(\bs_\blambda^2))$ where $\blambda = [\bmm_\blambda, \bs_\blambda]$, $\bmm_\blambda \in \mathbb{R}^D$, and $\bs_\blambda \in \mathbb{R}^D_+$. The sampling procedure could then be defined as
\begin{align}
	\epsilon &\sim \mathcal{N}(0, I_D) \, , \quad\quad 
	\bz = \mathcal{T}(\epsilon; \blambda) = \bmm_\blambda + \bs_\blambda \odot \epsilon
	\label{eq:gaussian-sample}
\end{align}
where $\bs \odot \epsilon$ denotes an element-wise product. We will also use $x/y$ and $x^2$ to denote pointwise division and squaring, respectively.
Given this map, the reparameterization gradient estimator is simply the gradient of a Monte Carlo ELBO estimate with respect to $\blambda$.
For a single sample, this is
\begin{align*}
	\epsilon &\sim q_0(\epsilon) \, , \quad\quad 
	\hat \bg_\blambda \triangleq \nabla_\blambda \left[ \ln p(\mathcal{T}(\epsilon;\blambda), \mathcal{D}) - \ln q(\mathcal{T}(\epsilon; \blambda); \blambda) \right]
\end{align*}
and the $L$-sample approximation can be computed by the sample average
\begin{align}
	\hat \bg_\blambda^{(L)} &= \frac{1}{L} \sum_{\ell=1}^L \hat \bg_\blambda(\epsilon^{\ell}).
\end{align}
Crucially, the reparameterization gradient is unbiased, $\mathbb{E}[\hat \bg_\blambda] = \nabla_\blambda \mathcal{L}(\blambda)$,
guaranteeing the convergence of stochastic gradient optimization procedures that use it~\cite{robbins1951stochastic}.

\label{sec:gradient-noise}
\begin{figure}[t!]
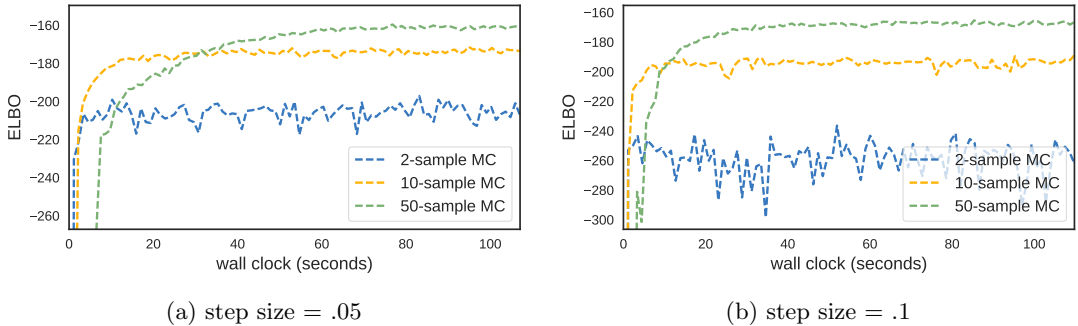

    \centering
    \begin{subfigure}[b]{.48\textwidth}
    	\centering
		\includegraphics[width=\textwidth]{{{figs/experiment_bnn/bnn_convergence-ss-0.050-mc-only}}}
        \caption{step size = .05}
        \label{fig:step-one}
    \end{subfigure}
	~
    \begin{subfigure}[b]{.48\textwidth}
    	\centering 
		\includegraphics[width=\textwidth]{{{figs/experiment_bnn/bnn_convergence-ss-0.100-mc-only}}}
        \caption{step size = .1}
        \label{fig:step-two}
    \end{subfigure}
\caption{Optimization traces for MCVI applied to a Bayesian neural network with various hyperparameter settings.  Each trace is running \texttt{adam} \cite{kingma2014adam}.  The three lines in each plot correspond to three different numbers of samples, $L$, used to estimate the gradient at each step.  (Left) small stepsize; (Right) stepsize 10 times larger.  Large step sizes allow for quicker progress, however noisier (i.e.,~small $L$) gradients combined with large step sizes result in chaotic optimization dynamics.  The converging traces reach different ELBOs due to the illustrative constant learning rates; in practice, one decreases the step size over time in keeping with \citet{robbins1951stochastic}.
}
\label{fig:grad-variance-example}
\vspace{-.5em}
\end{figure}

\vspace{-.5em}
\paragraph{Gradient Variance and Convergence}
The efficiency of Monte Carlo variational inference hinges on the magnitude of gradient noise and the step size chosen for the optimization procedure. 
When the gradient noise is large, smaller gradient steps must be taken to avoid unstable dynamics of the iterates.
However, a smaller step size increases the number of iterations that must be performed to reach convergence. 

We illustrate this trade-off in Figure~\ref{fig:grad-variance-example}, which shows realizations of an optimization procedure applied to a Bayesian neural network using reparameterization gradients.
The posterior is over~${D=653}$ parameters that we approximate with a diagonal Gaussian (see Appendix~\ref{sec:appendix-bnn}). 
We compare the progress of the \texttt{Adam} algorithm using various numbers of samples~\cite{kingma2014adam}, fixing the learning rate.
The noise present in the single-sample estimator causes extremely slow convergence, whereas the lower noise 50-sample estimator quickly converges, albeit at 50 times the cost.

The upshot is that with low noise gradients we are able to safely take larger steps, enabling faster convergence to a local optimum.
The natural question is, how can we reduce the variance of gradient estimates without introducing too much extra computation?
Our approach is to use information about the variational model, $q(\cdot; \blambda)$ and carefully construct a control variate to the gradient.
\label{sec:control-variates}
\vspace{-.5em}
\paragraph{Control Variates}
Control variates are random quantities that are used to reduce the variance of a statistical estimator without introducing any bias by injecting information into the estimator.
Given an unbiased estimator $\hat \bg$ such that $\mathbb{E}[\hat \bg] = \bg$ (the quantity of interest), our goal is to construct another unbiased estimator with lower variance.  
We can do this by defining a \emph{control variate} $\tilde \bg$ with \emph{known expectation} $\tilde \bmm$.
We can write our new estimator as 
\begin{align}
	\bg^{(cv)} &= \hat \bg - \bC(\tilde \bg - \tilde \bmm) \, .
	\label{eq:control-variate}
\end{align} 
where $\bC \in \mathbb{R}^{D \times D}$ for $D$-dimensional $\hat \bg$.
Clearly the new estimator has the same expectation as the original estimator, but a different variance.
We can attain optimal variance reduction by appropriately setting $\bC$.
Intuitively, the optimal $\bC$ is very similar to a regression coefficient --- it is related to the \textit{covariance} between the control variate and the original estimator. 
See Appendix~\ref{appendix-control-variates} for further details on optimally setting $\bC$.


\section{Method: Modeling Reparameterization Gradients}
\label{sec:methods}
In this section we develop our main contribution, a new gradient estimator that can dramatically reduce reparameterization gradient variance.
In MCVI, the reparameterization gradient estimator (RGE) is a Monte Carlo estimator of the true gradient --- the estimator itself is a random variable.
This random variable is generated using the ``reparameterization trick'' --- we first generate some randomness $\epsilon$ and then compute the gradient of the ELBO with respect to $\blambda$ holding $\epsilon$ fixed.
This results in a complex distribution from which we can generate samples, but in general cannot characterize due to the complexity of the term arising from the gradient of the true posterior. 

However, we do have a lot of information about the sampling procedure --- we know the variational distribution~$\ln q(\bz; \blambda)$, the transformation~$\mathcal{T}$, and we can evaluate the model joint density~$\ln p(\bz, \mathcal{D})$ pointwise.
Furthermore, with automatic differentiation, it is often straightforward to obtain gradients and Hessian-vector products of our model $\ln p(\bz, \mathcal{D})$.
We propose a scheme that uses the structure of $q_\blambda$ and curvature of $\ln p(\bz, \mathcal{D})$ to construct a tractable approximation of the distribution of the RGE.\footnote{We require the model $\ln p(\bz, \mathcal{D})$ to be twice differentiable.}
This approximation has a known mean and is correlated with the RGE distribution, allowing us to use it as a control variate to reduce the RGE variance. 

Given a variational family parameterized by $\blambda$, we can decompose the ELBO gradient into a few terms that reveal its ``data generating procedure''
\begin{align}
\epsilon \sim q_0 \, , \quad 
\bz &= \mathcal{T}(\epsilon; \blambda) \\
\hat \bg_\blambda \triangleq \hat \bg(\bz; \blambda)
 &= \underbrace{\frac{\partial \ln p(\bz, \mathcal{D})}{\partial \bz}}_{\text{\emph{data term}}} \frac{\partial \bz}{\partial \blambda} -
     \underbrace{\frac{\partial \ln q_\blambda(\bz)}{\partial \bz}}_{\text{\emph{pathwise score}}} \frac{\partial \bz}{\partial \blambda} - 
     \underbrace{\frac{\partial \ln q_\blambda(\bz)}{\partial \blambda}}_{\text{\emph{parameter score}}}.
     \label{eq:gradient}
\end{align}
Certain terms in Eq.~\eqref{eq:gradient} have tractable distributions.
The Jacobian of $\mathcal{T}(\cdot; \blambda)$, given by ${\partial \bz / \partial \blambda}$, is defined by our choice of $q(\bz; \blambda)$. 
For some transformations $\mathcal{T}$ we can exactly compute the distribution of the Jacobian given the distribution of $\epsilon$.
The \emph{pathwise} and \emph{parameter score} terms are gradients of our approximate distribution with respect to $\blambda$ (via $\bz$ or directly).
If our approximation is tractable (e.g.,~a multivariate Gaussian), we can exactly characterize the distribution for these components.\footnote{In fact, we know that the expectation of the \emph{parameter score} term is zero, and removing that term altogether can sometimes be a source of variance reduction that we do not explore here~\cite{roeder2017sticking}.}

However, the \emph{data term} in Eq.~\eqref{eq:gradient} involves a potentially complicated function of the latent variable $\bz$ (and therefore a complicated function of $\epsilon$), resulting in a difficult-to-characterize distribution. 
Our goal is to construct an approximation to the distribution of~${\partial \ln p(\bz, \mathcal{D}) / \partial \bz}$ and its interaction with~${\partial \bz / \partial \blambda}$ given a fixed distribution over $\epsilon$.
If the approximation yields random variables that are highly correlated with~$\hat \bg_\blambda$, then we can use it to reduce the variance of that RGE sample.
\vspace{-.5em}
\paragraph{Linearizing the data term}
To simplify notation, we write the data term of the gradient as 
\begin{align}
\bff(\bz')
  \triangleq \frac{\partial \ln p(\bz, \mathcal{D})}{\partial \bz} \Big |_{\bz=\bz'} \, \quad , \quad\quad
\end{align}
where~${\bff: \mathbb{R}^D \mapsto \mathbb{R}^D}$ since $\bz \in \mathbb{R}^D$.
We then linearize $\bff$ about some value $\bz_0$
\begin{align}
\tilde \bff(\bz) 
  &= \bff(\bz_0) + \left[ \bJ_{\bz} \bff(\bz)\right] (\bz - \bz_0) 
  = \bff(\bz_0) + \bH(\bz_0) (\bz - \bz_0) \
\label{eq:f-approx}
\end{align}  
where $\bH(\bz_0)$ is the Hessian of the model, $\ln p(\bz, \mathcal{D})$, with respect to $\bz$ evaluated at $\bz_0$,
\begin{align}
	\bH(\bz_0) &= \bJ_{\bz}\left[\frac{\partial \ln p(\bz, \mathcal{D})}{\partial \bz}\right]_{\bz = \bz_0} \, .
\end{align}
Note that even though this uses second-order information about the model, it is a first-order approximation of the gradient.
We also view this as a transformation of the random $\epsilon$ for a fixed $\blambda$
\begin{align}
\tilde \bff_\blambda(\epsilon) &= \bff(\bz_0) + \bH(\bz_0)( \mathcal{T}(\epsilon, \blambda) - \bz_0) \, ,
\label{eq:data-approx}
\end{align}
which is linear in~${\bz = \mathcal{T}(\epsilon, \blambda)}$.
For some forms of $\mathcal{T}$ we can analytically derive the distribution of the random variable $\tilde \bff_\blambda(\epsilon)$. 
In Eq.~\eqref{eq:gradient}, the \emph{data term} interacts with the Jacobian of $\mathcal{T}$, given by
\begin{align}
	\bJ_{\blambda'}(\epsilon) \triangleq \frac{\partial \bz}{\partial \blambda} 
	= \frac{\partial \mathcal{T}(\epsilon, \blambda)}{\partial \blambda}
	\Big |_{\blambda = \blambda'} \, .
\label{eq:jacobian}
\end{align}
which importantly is a function of the same $\epsilon$ as in Eq.~\eqref{eq:data-approx}.
We form our approximation of the first term in Eq.~\eqref{eq:gradient} by multiplying Eqs.~\eqref{eq:data-approx} and \eqref{eq:jacobian}:
\begin{align}
\tilde \bg_\blambda^{(data)}(\epsilon) &\triangleq \tilde \bff_\blambda(\epsilon) \bJ_\blambda(\epsilon)\,.
\label{eq:data-lam}
\end{align}  
The tractability of this approximation hinges on how Eq.~\eqref{eq:data-lam} depends on $\epsilon$.
When $q(\bz; \blambda)$ is multivariate normal, we show that this approximation has a computable mean and can be used to reduce variance in MCVI settings.
In the following sections we describe and empirically test this variance reduction technique applied to diagonal Gaussian posterior approximations.

\subsection{Gaussian Variational Families}
\label{sec:method-gaussian}
Perhaps the most common choice of approximating distribution for MCVI is a diagonal Gaussian, parameterized by a mean $\bmm_\blambda \in \mathbb{R}^D$ and scales $\bs_\blambda \in \mathbb{R}^D_+$. \footnote{For diagonal Gaussian $q$, we define $\blambda = [\bmm_\blambda, \bs_\blambda]$.}
The log pdf is 
\begin{align*}
\ln q(\bz; \bmm_\blambda, \bs_\blambda^2) 
= - \frac{1}{2}(\bz - \bmm_\blambda)^\intercal \left[\text{diag}(\bs_\blambda^2)\right]^{-1}(\bz - \bmm_\blambda)
 -\frac{1}{2}\sum_d \ln \bs_{\blambda,d}^2
 -\frac{D}{2}\ln(2\pi) \, .
\end{align*}
To generate a random variate $\bz$ from this distribution, we use the sampling procedure in Eq.~\eqref{eq:gaussian-sample}.
We denote the Monte Carlo RGE as ${\hat \bg_\blambda \triangleq [\hat \bg_{\bmm_\blambda}, \hat \bg_{\bs_\blambda}]}$. 
From this variational distribution, it is straightforward to derive the distributions of the \emph{pathwise score}, \emph{param score}, and \emph{Jacobian} terms in Eq.~\eqref{eq:gradient}. 

The \emph{Jacobian} term of the sampling procedure has two straightforward components
\begin{align}
	\frac{\partial \bz}{\partial \bmm_\blambda} = I_D \, , \quad 
	\frac{\partial \bz}{\partial \bs_\blambda} = \text{diag}(\epsilon) \, .
	\label{eq:transform-jacobian}
\end{align}

The \emph{pathwise score} term is the partial derivative of the approximate log density with respect to~$\bz$, ignoring variation due to the variational distribution parameters and noting that~${\bz = \bmm_\blambda + \bs_\blambda \odot \epsilon}$: 
\begin{align}
	\frac{\partial \ln q}{\partial \bz} &= - \text{diag}(\bs_\blambda^2)^{-1}(\bz - \bmm_\blambda)
	= - \epsilon / \bs_\blambda \,. 
	\label{eq:pathwise-score-jacobian}
\end{align}

The \emph{parameter score} term is the partial derivative of the approximation log density with respect to variational parameters $\blambda$, ignoring variation due to $\bz$.
The $\bmm_\blambda$ and $\bs_\blambda$ components are given by
\begin{align}
\label{eq:param-score-mean}
\frac{\partial \ln q}{\partial \bmm_\blambda} 
	&= (\bz - \bmm_\blambda) / \bs_\blambda^2
	= \epsilon/\bs_\blambda \, \\
\frac{\partial \ln q}{\partial s_\blambda} 
  &= -1/\bs_\blambda - (\bz - m_{\blambda})^2 / \bs_{\blambda}^2
  = \frac{\epsilon^2 - 1}{\bs_\blambda}.
 \label{eq:param-score-scale} 
\end{align}

The \emph{data term}, $\bff(\bz)$, multiplied by the Jacobian of $\mathcal{T}$ is all that remains to be approximated in Eq.~\eqref{eq:gradient}.
We linearize $\bff$ around $\bz_0 = \bmm_\blambda$ where the approximation is expected to be accurate
\begin{align}
	\tilde \bff_\blambda(\epsilon) &= \bff(\bmm_\blambda) + \bH(\bmm_\blambda)\left( (\bmm_\blambda + \bs_\blambda \odot \epsilon) - \bmm_\blambda \right) \\
	&\sim \mathcal{N}\left( \bff(\bmm_\blambda), \bH(\bmm_\blambda) \text{diag}(\bs_\blambda^2) \bH(\bmm_\blambda)^\intercal \right) \, .
	\label{eq:gauss-data-term}
\end{align}

\vspace{-.5em}
\paragraph{Putting It Together: Full RGE Approximation}
We write the complete approximation of the RGE in Eq.~\eqref{eq:gradient} by combining Eqs.~\eqref{eq:transform-jacobian}, \eqref{eq:pathwise-score-jacobian}, \eqref{eq:param-score-mean}, \eqref{eq:param-score-scale}, and \eqref{eq:gauss-data-term} which results in two components that are concatenated, $\tilde \bg_\blambda = [ \tilde \bg_{\bmm_\blambda}, \tilde \bg_{\bs_\blambda} ]$. Each component is defined as
\begin{align}
\tilde \bg_{\bmm_\blambda} 
	&= \tilde \bff_\blambda(\epsilon) + \epsilon / \bs_\blambda - \epsilon / \bs_\blambda 
	&&= \bff(\bmm_\blambda) + \bH(\bmm_\blambda)(\bs_\blambda \odot \epsilon) \label{eq:mean-approx-cv} \\
\tilde \bg_{\bs_\blambda} 
    &= \tilde \bff_\blambda(\epsilon) \odot \epsilon + (\epsilon / \bs_\blambda) \odot \epsilon -  \frac{\epsilon^2 - 1}{\bs_\blambda}
    &&= \left( \bff(\bmm_\blambda) + \bH(\bmm_\blambda)(\bs_\blambda \odot \epsilon) \right) \odot \epsilon +  \frac{1}{\bs_\blambda} \, . \label{eq:scale-approx-cv}
\end{align}

To summarize, we have constructed an approximation, $\tilde \bg_\blambda$, of the reparameterization gradient, $\hat \bg_\blambda$, as a function of $\epsilon$.
Because both $\tilde \bg_\blambda$ and $\hat \bg_\blambda$ are functions of the same random variable $\epsilon$, and because we have mimicked the random process that generates true gradient samples, the two gradient estimators will be correlated.
This approximation yields two tractable distributions --- a Gaussian for the mean parameter gradient,~$\bg_{\bmm_\blambda}$, and a location shifted, scaled non-central $\chi^2$ for the scale parameter gradient $\bg_{\bs_\blambda}$.
Importantly, we can compute the mean of each component
\begin{align}
	\mathbb{E}[\tilde \bg_{\bmm_{\blambda}} ] = \bff(\bmm_\blambda) \, , \quad\quad 
	\mathbb{E}[\tilde \bg_{\bs_\blambda} ] = \text{diag}(\bH(\bmm_\blambda)) \odot \bs_\blambda + 1 / \bs_\blambda \, .
	\label{eq:cv-mean}
\end{align}
We use $\tilde \bg_\blambda$ (along with its expectation) as a control variate to reduce the variance of the RGE $\hat \bg_\blambda$.

\subsection{Reduced Variance Reparameterization Gradient Estimators}

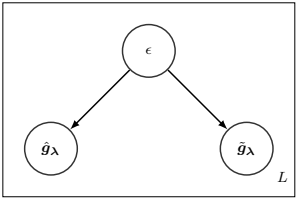
\begin{figure}[t!]
\begin{minipage}{.28\textwidth}
	\begin{figure}[H]
	  \centering
      \scalebox{.7}{\begin{tikzpicture}
\tikzstyle{main}=[circle, minimum size = 10mm, thick, draw =black!80, node distance = 16mm]
\tikzstyle{connect}=[-latex, thick]
\tikzstyle{box}=[rectangle, draw=black!100]
  \node[main] (eps) { $\epsilon$ };
  \node[main] (hatg) [below left=of eps] { $\hat \bg_\blambda$ };
  \node[main] (tildeg) [below right=of eps]{ $\tilde \bg_\blambda$ };
  \path (eps) edge [connect] (hatg)
        (eps) edge [connect] (tildeg);
  \node[rectangle, inner sep=4mm, draw=black!100, fit=(eps) (hatg) (tildeg)] (plate) {};
  \node (B) [below=of plate.east, rectangle] {}node at ($(B.south east) +(-0.35,-0.25)$) {$L$};
\end{tikzpicture}}
      \caption{Relationship between the base randomness $\epsilon$, RGE $\hat \bg$, and approximation $\tilde \bg$.
Arrows indicate deterministic functions.
Sharing $\epsilon$ correlates the random variables.
We know the distribution of $\tilde \bg$, which allows us to use it as a control variate for $\hat \bg$.}
	  \label{fig:graph-model}
	\end{figure}
\end{minipage}
~\quad
\begin{minipage}{0.68\textwidth}
	\begin{algorithm}[H]\scriptsize
	\caption{Gradient descent with RV-RGE with a diagonal Gaussian variational family} 
	\label{alg:reduced-variance}
	\begin{algorithmic}[1]
\Procedure{RV-RGE-Optimize}{$\blambda_1,\ln p(\bz, \mathcal{D}),\L$}
\State $\bff(\bz) \gets \nabla_{\bz} \ln p(\bz, \mathcal{D})$
\State $\bH(\bz_a, \bz_b) \gets \left[ \nabla_{\bz}^2 \ln p(\bz_a, \mathcal{D})\right] \bz_b$
\For{$t = 1, \dots, T$}
\State $\epsilon^{(\ell)} \sim \mathcal{N}(0, I_D) \text{ for } \ell = 1, \dots, L$ \Comment{Base randomness $q_0$}
\State $\hat \bg_{\blambda_t}^{(\ell)} \gets \nabla_{\blambda} \ln p(\bz(\epsilon^{(\ell)}, \blambda_t), \mathcal{D})$ \Comment{ Reparameterization gradients} 
\State $\tilde \bg_{\bmm_{\blambda_t}}^{(\ell)} \gets \bff(\bmm_{\blambda_t}) + \bH(\bmm_{\blambda_t}, \bs_{\blambda_t} \odot \epsilon^{(\ell)})$  \Comment{ Mean approx }
\State $\tilde \bg_{\bs_{\blambda_t}}^{(\ell)} \gets \left( \bff(\bmm_{\blambda_t}) + \bH(\bmm_{\blambda_t}, \bs_{\blambda_t} \odot \epsilon^{(\ell)}) \right) \odot \epsilon +  \frac{1}{\bs_{\blambda_t}}$ \Comment{ Scale approx }
\State $\mathbb{E}[\tilde \bg_{\bmm_{\blambda_t}} ] \gets \bff(\bmm_{\blambda_t})$ \Comment{ Mean approx expectation}
\State $\mathbb{E}[\tilde \bg_{\bs_{\blambda_t}} ] \gets \text{diag}(\bH(\bmm_{\blambda_t})) \odot \bs_{\blambda_t} + 1 / \bs_{\blambda_t}$ \Comment{ Scale approx expectation}
\State $\hat \bg_{\blambda_t}^{(RV)} = \frac{1}{L} \sum_\ell \hat \bg_{\blambda_t}^\ell - (\tilde \bg_{\blambda_t}^\ell - \mathbb{E}[\tilde \bg_{\blambda_t}] )$ \Comment{ Subtract control variate }
\State $\blambda_{t+1} \gets \texttt{grad-update}(\blambda_t, \hat \bg_{\blambda_t}^{(RV)})$ \Comment{Gradient step (\texttt{sgd}, \texttt{adam}, etc.)}
\EndFor
\State \textbf{return} $\blambda_T$
\EndProcedure
\end{algorithmic}
	\end{algorithm}    
\end{minipage}
\vspace{-1em}
\end{figure}

Now that we have constructed a tractable gradient approximation, $\tilde \bg_\blambda$, with high correlation to the original reparameterization gradient estimator, $\hat \bg_\blambda$, we can use it as a control variate as in
Eq.~\eqref{eq:control-variate}
\begin{align}
	\hat \bg_\blambda^{(RV)} &= \hat \bg_\blambda - \bC (\tilde \bg_\blambda - \mathbb{E}[\tilde \bg_\blambda] ).
\end{align}
The optimal value for $\bC$ is the covariance between $\tilde \bg_\blambda$ and $\hat \bg_\blambda$ (see Appendix~\ref{appendix-control-variates}).
We can try to estimate the value of $\bC$ (or a diagonal approximation to $\bC$) on the fly, or we can simply fix this value. 
In our case, because we are using an accurate linear approximation to the transformation of a spherical Gaussian, the optimal value of $\bC$ will be close to the identity (see Appendix~\ref{appendix-optimal-c}).

\vspace{-.5em}
\paragraph{High Dimensional Models}
For models with high dimensional posteriors, direct manipulation of the Hessian is computationally intractable.
However, our approximations in Eqs.~\eqref{eq:mean-approx-cv} and~\eqref{eq:scale-approx-cv} only require a Hessian-vector product, which can be computed nearly as efficiently as the gradient~\cite{pearlmutter1994fast}. 
We note that the mean of the control variate $\tilde \bg_{\bs_\blambda}$ (Eq.~\eqref{eq:cv-mean}), depends on the diagonal of the Hessian matrix. 
While computing the Hessian diagonal may be tractable in some cases, in general it may cost the time equivalent of $D$ function evaluations to compute \cite{martens2012estimating}. 
Given a high dimensional problem, we can avoid this bottleneck in multiple ways.
The first is simply to ignore the random variation in the Jacobian term due to $\epsilon$ --- if we fix $\bz$ to be $\bmm_\blambda$ (as we do with the data term), the portion of the Jacobian that corresponds to $\bs_\blambda$ will be zero (in Eq.~\eqref{eq:transform-jacobian}).
This will result in the same Hessian-vector-product-based estimator for $\tilde \bg_{\bmm_\blambda}$ but will set ${\tilde \bg_{\bs_\blambda} = 0}$, yielding variance reduction for the mean parameter but not the scale.

Alternatively, we can estimate the Hessian diagonal on the fly.
If we use $L > 1$ samples at each iteration, we can create a per-sample estimate of the $\bs_\blambda$-scaled diagonal of the Hessian using the other samples \cite{bekas2007estimator}.
As the scaled diagonal estimator is unbiased, we can construct an unbiased estimate of the control variate mean to use in lieu of the actual mean (possibly increasing the final variance). 
A similar \emph{local baseline} strategy is used for variance reduction in \citet{mnih2016variational}.

\vspace{-.5em}
\paragraph{RV-RGE Estimators}
We introduce three different estimators based on variations of the gradient approximation defined in Eqs.~\eqref{eq:mean-approx-cv}, \eqref{eq:scale-approx-cv}, and \eqref{eq:cv-mean}, each adressing the Hessian operations differently. 
\vspace{\parvspace}
\begin{itemize} \itemsep 0pt
\item The \emph{Full Hessian} estimator implements the three equations as written and can be used when it is computationally feasible to use the full Hessian.
\item The \emph{Hessian Diagonal} estimator replaces the Hessian in \eqref{eq:mean-approx-cv} with a diagonal approximation, useful for models with a cheap Hessian diagonal.
\item The \emph{Hessian-vector product + local approximation} (HVP+Local) uses an efficient Hessian-vector product in Eqs.~\eqref{eq:mean-approx-cv} and \eqref{eq:scale-approx-cv}, while approximating the diagonal term in Eq.~\eqref{eq:cv-mean} using a local baseline.  The HVP+Local approximation is geared toward models where Hessian-vector products can be computed, but the exact diagonal of the Hessian cannot.
\end{itemize}
\vspace{\parvspace}
We detail the RV-RGE algorithm in Algorithm~\ref{alg:reduced-variance} and
compare properties of these three estimators to the pure Monte Carlo estimator in the following section.

\subsection{Related Work}
\label{sec:related-work}

Recently, \citet{roeder2017sticking} introduced a variance reduction technique for reparameterization gradients that ignores the \emph{parameter score} component of the gradient and can be viewed as a type of control variate for the gradient throughout the optimization procedure.
This approach is complementary to our method --- our approximation is typically more accurate near the beginning of the optimization procedure, whereas the estimator in \citet{roeder2017sticking} is low-variance near convergence. 
We hope to incorporate information from both control variates in future work. 
Per-sample estimators in a multi-sample setting for variational inference were used in \citet{mnih2016variational}.
We employ this technique in a different way; we use it to estimate computationally intractable quantities needed to keep the gradient estimator unbiased.
Black box variational inference used control variates and Rao-Blackwellization to reduce the variance of score-function estimators~\cite{ranganath2014black}.  
Our development of variance reduction for reparameterization gradients compliments their work. 
Other variance reduction techniques for stochastic gradient descent have focused on stochasticity due to data subsampling \cite{johnson2013accelerating, wang2013variance}.
\citet{johnson2013accelerating} cache statistics about the entire dataset at each epoch to use as a control variate for noisy mini-batch gradients.

\section{Experiments and Analysis}
\label{sec:experiments}

\begin{table}[t!]
  \centering
  \caption{Comparing variances for RV-RGEs: $L=10$-sample estimators. Values are percentage of pure MC RGE variance --- a value of 100 indicates equal variation $L=10$ samples, a value of 1 percent indicates a 100-fold decrease in variance (lower is better).}
  \label{table:variance}
  \scalebox{.9}{
  
\begin{tabular}{l l r r r r r r}
  \toprule
            &              & \multicolumn{2}{c}{$\bg_{\bmm_\blambda}$} & \multicolumn{2}{c}{$\ln \bg_{\bs_\blambda}$} & \multicolumn{2}{c}{$\bg_{\blambda}$} \\
  Iteration & Estimator    & Ave $\mathbb{V}(\cdot)$ & $\mathbb{V}( ||\cdot||)$ & Ave $\mathbb{V}(\cdot)$ & $\mathbb{V}( ||\cdot||)$ & Ave $\mathbb{V}(\cdot)$ & $\mathbb{V}( ||\cdot||)$ \\
  \midrule
  
  \multirow{4}{*}{ early }
    & Full Hessian & 1.279 & 1.139 & 0.001 & 0.002 & 0.008 & 1.039 \\
    & Hessian Diag & 34.691 & 23.764 & 0.003 & 0.012 & 0.194 & 21.684 \\
    & HVP + Local  & 1.279 & 1.139 & 0.013 & 0.039 & 0.020 & 1.037 \\
  \midrule

  \multirow{4}{*}{ mid }
    & Full Hessian & 0.075 & 0.068 & 0.113 & 0.143 & 0.076 & 0.068 \\
    & Hessian Diag & 38.891 & 21.283 & 6.295 & 7.480 & 38.740 & 21.260 \\
    & HVP + Local  & 0.075 & 0.068 & 30.754 & 39.156 & 0.218 & 0.071 \\
  \midrule

  \multirow{4}{*}{ late }
    & Full Hessian & 0.042 & 0.030 & 1.686 & 0.431 & 0.043 & 0.030 \\
    & Hessian Diag & 40.292 & 53.922 & 23.644 & 28.024 & 40.281 & 53.777 \\
    & HVP + Local  & 0.042 & 0.030 & 98.523 & 99.811 & 0.110 & 0.022 \\
  \bottomrule

\end{tabular}

  }
  \vspace{-1em}
\end{table}

\begin{figure*}[t!]
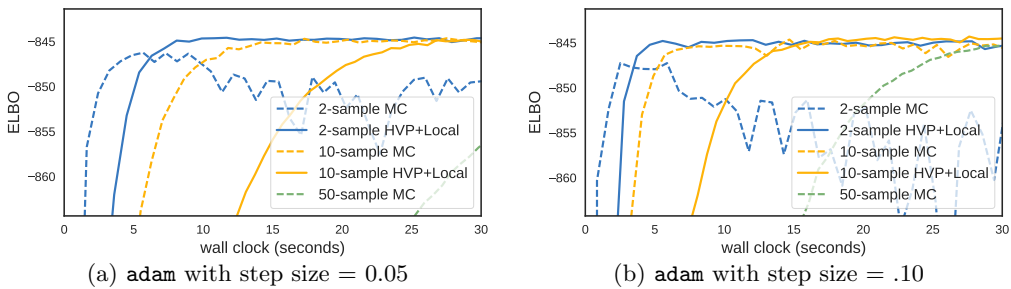

    \centering
    \begin{subfigure}[b]{.45\columnwidth}
    	\centering
    	\includegraphics[width=\columnwidth]{{{figs/experiment_frisk/frisk_convergence-ss-0.050-with-50}}}
  		\vspace{-0.6cm}%
    	\caption{\texttt{adam} with step size = 0.05}
        \label{fig:step-one}
    \end{subfigure}
    ~
    \begin{subfigure}[b]{.45\columnwidth}
    	\centering 
    	\includegraphics[width=\columnwidth]{{{figs/experiment_frisk/frisk_convergence-ss-0.100-with-50}}}
  		\vspace{-0.6cm}%
    	\caption{\texttt{adam} with step size = .10}
        \label{fig:step-two}
    \end{subfigure}
    \vspace{-0.1cm}%
\caption{MCVI optimization trace applied to the \texttt{frisk} model for two values of $L$ and step size.  We run the standard MC gradient estimator (solid line) and the RV-RGE with ${L=2}$ and $10$ samples.}
\vspace{-1em}
\label{fig:frisk}
\end{figure*}

\begin{figure}[t!]
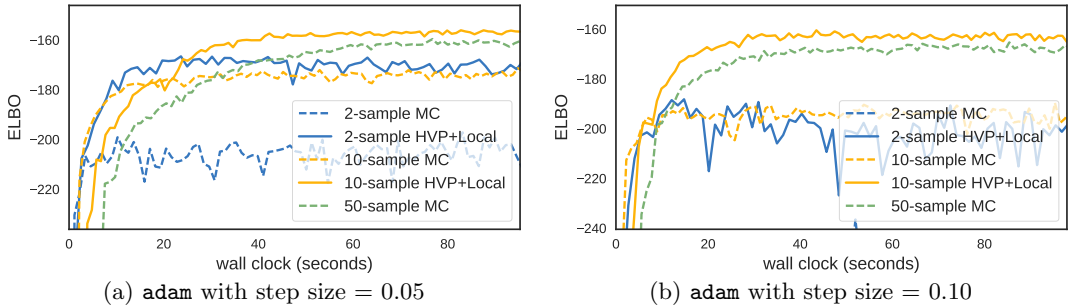

	\centering
    \begin{subfigure}[b]{.48\columnwidth}
    	\centering
  		\includegraphics[width=\columnwidth]{{{figs/experiment_bnn/bnn_convergence-ss-0.050-with-50}}}
  		\vspace{-0.6cm}%
        \caption{\texttt{adam} with step size = 0.05}
        \label{fig:bnn-one}
    \end{subfigure}~
    \begin{subfigure}[b]{.48\columnwidth}
    	\centering
  		\includegraphics[width=\columnwidth]{{{figs/experiment_bnn/bnn_convergence-ss-0.100-with-50}}}
  		\vspace{-0.6cm}%
        \caption{\texttt{adam} with step size = 0.10}
        \label{fig:bnn-two}
    \end{subfigure}
    \vspace{-0.1cm}%
\caption{MCVI optimization for the \texttt{bnn} model applied to the \texttt{wine} data for various $L$ and step sizes. The standard MC gradient estimator (dotted) was run with 2, 10, and 50 samples; RV-RGE (solid) was run with 2 and 10 samples.
In \ref{fig:bnn-two} the 2-sample MC estimator falls below the frame. 
}
\vspace{-1em}
\label{fig:bnn}
\end{figure}

In this section we empirically examine the variance properties of RVRGs and stochastic optimization for two real-data examples --- a hierarchical Poisson GLM and a Bayesian neural network.\footnote{Code is available at \url{https://github.com/andymiller/ReducedVarianceReparamGradients}.}
\vspace{\parvspace}
\begin{itemize} \itemsep 0pt

\item \emph{Hierarchical Poisson GLM}
The \texttt{frisk} model is a hierarchical Poisson GLM, described in Appendix~\ref{sec:appendix-frisk}. 
This non-conjugate model has a $D=37$ dimensional posterior.

\item \emph{Bayesian Neural Network}
The non-conjugate \texttt{bnn} model is a Bayesian neural network applied to the \texttt{wine} dataset, (see Appendix~\ref{sec:appendix-bnn}) and has a $D=653$ dimensional posterior. 
\end{itemize}

\vspace{-.5em}
\paragraph{Quantifying Gradient Variance Reduction}
We measure the variance reduction of the RGE observed at various iterates, $\blambda_t$, during execution of gradient descent.
Both the gradient magnitude, and the marginal variance of the gradient elements --- using a sample of 1000 gradients --- are reported.
Further, we inspect both the mean $\bmm_\blambda$ and log-scale $\ln \bs_\blambda$ parameters separately. 
Table~\ref{table:variance} compares gradient variances for the \texttt{frisk} model for four estimators: i) pure Monte Carlo (MC), ii) Full Hessian, iii) Hessian Diagonal, and iv) Hessian-vector product + local approximation (HVP+Local).

Each entry in the table measures the percent of the variance of the pure Monte Carlo estimator.  
We show the average variance over each component $\text{Ave} \mathbb{V}(\cdot)$, and the variance of the norm $\mathbb{V}(|| \cdot ||)$.
We separate out variance in mean parameters, $\bg_{\bmm}$, log scale parameters, $\ln \bg_{\bs}$, and the entire vector $\bg_\blambda$.
The reduction in variance is dramatic.
Using HVP+Local, in the norm of the mean parameters we see between a $80 \times$ and $3{,}000 \times$ reduction in variance depending on the progress of the optimizer.
The importance of the full Hessian-vector product for reducing mean parameter variance is also demonstrated as the Hessian diagonal only reduces mean parameter variance by a factor of 2-5.

For the variational scale parameters, $\ln \bg_{\bs}$, we see that early on the HVP+Local approximation is able to reduce parameter variance by a large factor ($\approx 2{,}000 \times$).
However, at later iterates the HVP+Local scale parameter variance is on par with the Monte Carlo estimator, while the full Hessian estimator still enjoys huge variance reduction. 
This indicates that, by this point, most of the noise is the local Hessian diagonal estimator.
We also note that in this problem, most of the estimator variance is in the mean parameters.
Because of this, the norm of the entire parameter gradient, $\bg_\blambda$ is reduced by $100-5{,}000 \times$.
In Appendix~\ref{appendix-variance} we report results for other values of $L$ as a comparison.

\vspace{-.5em}
\paragraph{Optimizer Convergence and Stability}
We compare the optimization traces for the \texttt{frisk} and \texttt{bnn} model for the MC and the HVP+Local estimators under various conditions.
At each iteration we estimate the true ELBO value using 2000 Monte Carlo samples.
We optimize the ELBO objective using \texttt{adam} \cite{kingma2014adam} for two step sizes, each trace starting at the same value of $\blambda_0$.

Figure~\ref{fig:frisk} compares ELBO optimization traces for~${L=2}$ and~${L=10}$ samples and step sizes~$.05$ and~$.1$ for the \texttt{frisk} model.
We see that the HVP+Local estimators make early progress and converge quickly.
We also see that the~${L=3}$ pure MC estimator results in noisy optimization paths. 
Figure~\ref{fig:bnn} shows objective value as a function of wall clock time under various settings for the \texttt{bnn} model.
The HVP+Local estimator does more work per iteration, however it tends to converge faster. 
We observe the ${L=10}$ HVP+Local outperforming the ${L=50}$ MC estimator.

\section{Conclusion}
\label{sec:conclusion}
Variational inference reframes an integration problem as an optimization
problem with the caveat that each step of the optimization procedure solves an
easier integration problem.
For general models, each sub-integration problem is itself intractable, and
must be estimated, typically with Monte Carlo samples.
Our work has shown that we can use more information about the variational
family to create tighter estimators of the ELBO gradient, which leads to faster and
more stable optimization. The efficacy of our approach relies on the complexity of the RGE distribution to be well-captured by linear structure which may not be true for all models. However, we found the idea effective for non-conjugate hierarchical Bayesian models and a neural network.

Our presentation is a specific instantiation of a more general idea --- using
cheap linear structure to remove variation from stochastic gradient estimates.
We would like to extend this idea to more flexible variational distributions,
including flow distributions \cite{rezende2015variational} and hierarchical
distributions \cite{ranganath2016hierarchical}, as well as model/inference
schemes with recognition networks \cite{kingma2013auto}.

\subsubsection*{Acknowledgements}
The authors would like to thank Finale Doshi-Velez, Mike Hughes, Taylor Killian, Andrew Ross, and Matt Hoffman for helpful conversations and comments on this work.
ACM is supported by the Applied Mathematics Program within the Office of Science Advanced Scientific Computing Research of the U.S. Department of Energy under contract No. DE-AC02-05CH11231.
NJF is supported by a Washington Research Foundation Innovation Postdoctoral Fellowship in Neuroengineering and Data Science.
RPA is supported by NSF IIS-1421780 and the Alfred P. Sloan Foundation.

\small
\bibliographystyle{plainnat}
\bibliography{refs.bib}

\begin{thebibliography}{24}
\providecommand{\natexlab}[1]{#1}
\providecommand{\url}[1]{\texttt{#1}}
\expandafter\ifx\csname urlstyle\endcsname\relax
  \providecommand{\doi}[1]{doi: #1}\else
  \providecommand{\doi}{doi: \begingroup \urlstyle{rm}\Url}\fi

\bibitem[Arjovsky et~al.(2017)Arjovsky, Chintala, and
  Bottou]{arjovsky2017wasserstein}
Martin Arjovsky, Soumith Chintala, and L{\'e}on Bottou.
\newblock Wasserstein {GAN}.
\newblock \emph{arXiv preprint arXiv:1701.07875}, 2017.

\bibitem[Bekas et~al.(2007)Bekas, Kokiopoulou, and Saad]{bekas2007estimator}
Costas Bekas, Effrosyni Kokiopoulou, and Yousef Saad.
\newblock An estimator for the diagonal of a matrix.
\newblock \emph{Applied numerical mathematics}, 57\penalty0 (11):\penalty0
  1214--1229, 2007.

\bibitem[Blei et~al.(2017)Blei, Kucukelbir, and McAuliffe]{blei2016variational}
David~M Blei, Alp Kucukelbir, and Jon~D McAuliffe.
\newblock Variational inference: A review for statisticians.
\newblock \emph{Journal of the American Statistical Association}, 2017.

\bibitem[Gelman and Hill(2006)]{gelman2006data}
Andrew Gelman and Jennifer Hill.
\newblock \emph{Data Analysis Using Regression and Multilevel/Hierarchical
  Models}.
\newblock Cambridge University Press, 2006.

\bibitem[Gelman et~al.(2007)Gelman, Fagan, and Kiss]{gelman2007analysis}
Andrew Gelman, Jeffrey Fagan, and Alex Kiss.
\newblock An analysis of the {NYPD}’s stop-and-frisk policy in the context of
  claims of racial bias.
\newblock \emph{Journal of the American Statistical Association}, 102:\penalty0
  813--823, 2007.

\bibitem[Goodfellow et~al.(2014)Goodfellow, Pouget-Abadie, Mirza, Xu,
  Warde-Farley, Ozair, Courville, and Bengio]{goodfellow2014generative}
Ian Goodfellow, Jean Pouget-Abadie, Mehdi Mirza, Bing Xu, David Warde-Farley,
  Sherjil Ozair, Aaron Courville, and Yoshua Bengio.
\newblock Generative {A}dversarial {N}ets.
\newblock In \emph{Advances in Neural Information Processing Systems}, pages
  2672--2680, 2014.

\bibitem[Hoffman et~al.(2013)Hoffman, Blei, Wang, and
  Paisley]{hoffman2013stochastic}
Matthew~D Hoffman, David~M Blei, Chong Wang, and John~William Paisley.
\newblock Stochastic variational inference.
\newblock \emph{Journal of Machine Learning Research}, 14\penalty0
  (1):\penalty0 1303--1347, 2013.

\bibitem[Johnson and Zhang(2013)]{johnson2013accelerating}
Rie Johnson and Tong Zhang.
\newblock Accelerating stochastic gradient descent using predictive variance
  reduction.
\newblock In \emph{Advances in Neural Information Processing Systems}, pages
  315--323, 2013.

\bibitem[Kingma and Ba(2015)]{kingma2014adam}
Diederik Kingma and Jimmy Ba.
\newblock Adam: A method for stochastic optimization.
\newblock In \emph{Proceedings of the International Conference on Learning
  Representations}, 2015.

\bibitem[Kingma and Welling(2014)]{kingma2013auto}
Diederik~P Kingma and Max Welling.
\newblock Auto-encoding variational {B}ayes.
\newblock In \emph{Proceedings of the International Conference on Learning
  Representations}, 2014.

\bibitem[Maclaurin et~al.(2015)Maclaurin, Duvenaud, Johnson, and
  Adams]{autograd}
Dougal Maclaurin, David Duvenaud, Matthew Johnson, and Ryan~P. Adams.
\newblock Autograd: Reverse-mode differentiation of native {P}ython, 2015.
\newblock URL \url{http://github.com/HIPS/autograd}.

\bibitem[Martens et~al.(2012)Martens, Sutskever, and
  Swersky]{martens2012estimating}
James Martens, Ilya Sutskever, and Kevin Swersky.
\newblock Estimating the {H}essian by back-propagating curvature.
\newblock In \emph{Proceedings of the International Conference on Machine
  Learning}, 2012.

\bibitem[Mnih and Rezende(2016)]{mnih2016variational}
Andriy Mnih and Danilo Rezende.
\newblock Variational inference for {M}onte {C}arlo objectives.
\newblock In \emph{Proceedings of The 33rd International Conference on Machine
  Learning}, pages 2188--2196, 2016.

\bibitem[Mohamed and Lakshminarayanan(2016)]{mohamed2016learning}
Shakir Mohamed and Balaji Lakshminarayanan.
\newblock Learning in implicit generative models.
\newblock \emph{arXiv preprint arXiv:1610.03483}, 2016.

\bibitem[Pearlmutter(1994)]{pearlmutter1994fast}
Barak~A Pearlmutter.
\newblock Fast exact multiplication by the {H}essian.
\newblock \emph{Neural computation}, 6\penalty0 (1):\penalty0 147--160, 1994.

\bibitem[Ranganath et~al.(2014)Ranganath, Gerrish, and
  Blei]{ranganath2014black}
Rajesh Ranganath, Sean Gerrish, and David~M Blei.
\newblock Black box variational inference.
\newblock In \emph{AISTATS}, pages 814--822, 2014.

\bibitem[Ranganath et~al.(2016)Ranganath, Tran, and
  Blei]{ranganath2016hierarchical}
Rajesh Ranganath, Dustin Tran, and David~M Blei.
\newblock Hierarchical variational models.
\newblock In \emph{International Conference on Machine Learning}, 2016.

\bibitem[Rezende and Mohamed(2015)]{rezende2015variational}
Danilo Rezende and Shakir Mohamed.
\newblock Variational inference with normalizing flows.
\newblock In \emph{Proceedings of the 32nd International Conference on Machine
  Learning (ICML-15)}, pages 1530--1538, 2015.

\bibitem[Rezende et~al.(2014)Rezende, Mohamed, and
  Wierstra]{rezende2014stochastic}
Danilo~Jimenez Rezende, Shakir Mohamed, and Daan Wierstra.
\newblock Stochastic backpropagation and approximate inference in deep
  generative models.
\newblock In \emph{International Conference on Machine Learning}, 2014.

\bibitem[Robbins and Monro(1951)]{robbins1951stochastic}
Herbert Robbins and Sutton Monro.
\newblock A stochastic approximation method.
\newblock \emph{The Annals of Mathematical Statistics}, pages 400--407, 1951.

\bibitem[Roeder et~al.(2017)Roeder, Wu, and Duvenaud]{roeder2017sticking}
Geoffrey Roeder, Yuhuai~Wu Wu, and David Duvenaud.
\newblock Sticking the landing: An asymptotically zero-variance gradient
  estimator for variational inference.
\newblock \emph{arXiv preprint arXiv:1703.09194}, 2017.

\bibitem[Ruiz et~al.(2016)Ruiz, AUEB, and Blei]{ruiz2016generalized}
Francisco~R Ruiz, Michalis Titsias~RC AUEB, and David Blei.
\newblock The generalized reparameterization gradient.
\newblock In \emph{Advances in Neural Information Processing Systems}, pages
  460--468, 2016.

\bibitem[Tran et~al.(2017)Tran, Hoffman, Saurous, Brevdo, Murphy, and
  Blei]{tran2017deep}
Dustin Tran, Matthew~D Hoffman, Rif~A Saurous, Eugene Brevdo, Kevin Murphy, and
  David~M Blei.
\newblock Deep probabilistic programming.
\newblock In \emph{Proceedings of the International Conference on Learning
  Representations}, 2017.

\bibitem[Wang et~al.(2013)Wang, Chen, Smola, and Xing]{wang2013variance}
Chong Wang, Xi~Chen, Alexander~J Smola, and Eric~P Xing.
\newblock Variance reduction for stochastic gradient optimization.
\newblock In \emph{Advances in Neural Information Processing Systems}, pages
  181--189, 2013.

\end{thebibliography}

\clearpage

\appendix
\section{Control Variates}
\label{appendix-control-variates}
Control variates are random quantities that are used to reduce the variance of a statistical estimator without trading any bias.  
Concretely, given an unbiased estimator $\hat \bg$ such that $\mathbb{E}[\hat \bg] = \bg$ (the quantity of interest), our goal is to construct another unbiased estimator with lower variance.  
We can do this by defining a \emph{control variate} $\tilde \bg$ with \emph{known expectation} $\tilde \bmm$.
We can write our new estimator as 
\begin{align}
	\bg^{(cv)} &= \hat \bg - c \cdot(\tilde \bg - \tilde \bmm) \, .
\end{align} 
Clearly the new estimator has the same expectation as the original estimator, but a different variance.
We can reduce the variance of $\bg^{(cv)}$ by setting $c$ optimally.  

Consider a univariate $\hat \bg$ and $\tilde \bg$, and without loss of generality, take $\tilde \bmm = 0$.  
The variance of $\bg^{(cv)}$ can be written
\begin{align}
	\mathbb{V}(\bg^{(cv)}) 
	&= \mathbb{E}[(\hat \bg - c \cdot \tilde \bg)^2 ] - \mathbb{E}[\hat \bg]^2 \\
	&= \mathbb{E}[\hat \bg^2 + c^2 \cdot \tilde \bg^2 - 2 c \hat \bg \tilde \bg ] - \mathbb{E}[\hat \bg]^2\\
	&= \mathbb{E}[\hat \bg^2] + c^2 \mathbb{E}[\tilde \bg^2] - 2 c \mathbb{E}[\hat \bg \tilde \bg]- \mathbb{E}[\hat \bg]^2
\end{align}
We minimize the variance with respect to $c$ by taking the derivative and setting equal to zero, which implies
\begin{align}
	c^* = \frac{\mathbb{E}[\hat \bg \tilde \bg]}{\mathbb{E}[\tilde \bg^2]} 
	    = \frac{\mathbb{C}(\hat \bg, \tilde \bg)}{\mathbb{V}(\tilde \bg)}
\end{align}
The covariance $\mathbb{C}(\hat \bg, \tilde \bg)$ is typically not known a priori and must be estimated.  
It can be shown, under the optimal $c^*$, that the variance of $\bg^{(cv)}$ is 
\begin{align}
	\mathbb{V}(\bg^{(cv)}) &= (1 - \rho^2) \mathbb{V}(\hat \bg)
\end{align}
where $\rho$ is the correlation coefficient between $\tilde \bg$ and $\hat \bg$.

When $\hat \bg$ and $\tilde \bg$ are length $D$ vectors, we can construct an estimator that depends on a matrix-valued free parameter, $\bC \in \mathbb{R}^{D \times D}$
\begin{align}
	\bg^{(cv)} &= \hat \bg - \bC (\tilde \bg - \tilde \bmm) \, .
\end{align}
We can show that the $\bC$ that minimizes the $\texttt{Tr}(\mathbb{C}(\bg^{(cv)}))$ --- the sum of the marginal variances --- is given by 
\begin{align}
	\bC^* &= \Sigma_{\tilde \bg}^{-1} \Sigma_{\hat \bg, \tilde \bg}
\end{align}
where $\Sigma_{\tilde \bg}$ is the covariance matrix of the control variate vector, and $\Sigma_{\hat \bg, \tilde \bg}$ is the cross covariance between $\hat \bg$ and $\tilde \bg$.  

Intuitively, a control variate is injecting information into the estimator in the form of linear structure.
If the two quantities, $\tilde \bg$ and $\hat \bg$ are perfectly correlated, then we already know the mean and estimation is not necessary.
As the two become uncorrelated, the linear estimator becomes less and less informative, and reverts to the original quantity.

\subsection{Control Variates and Approximate Functions}
\label{appendix-optimal-c}
In our setting, we approximate the distribution of some function $\bff(\epsilon)$ where $\epsilon \sim \mathcal{N}(0, I)$ by a first order Taylor expansion about $0$ --- for now we examine the univariate case
\begin{align}
	\bff_1(\epsilon) = \bff(0) + \bff'(0) \epsilon \, \quad \epsilon \in \mathbb{R}
\end{align}

If we wish to use $\bff_1(\epsilon)$ as a control variate for $\bff(\epsilon)$, we need to characterize the covariance between the two random variables.
Because the form of $\bff(\epsilon)$ is general, it is difficult to analyze.
We instead derive the covariance between $\bff_1(\epsilon)$ and the second-order expansion
\begin{align}
	\bff_2(\epsilon) = \bff(0) + \bff'(0)\epsilon + \bff''(0)/2 \epsilon^2
\end{align}
as a surrogate. 

\begin{align}
\mathbb{C}(\bff_1(\epsilon), \bff_2(\epsilon))
	 &= \mathbb{E}\left[ (\bff_1(\epsilon) - \mathbb{E}[\bff_1(\epsilon)]) (\bff_2(\epsilon) - \mathbb{E}[\bff_2(\epsilon)]) \right] \\
	 &= \mathbb{E}\left[ (\bff'(0) \epsilon) \left(\bff'(0) \epsilon + \bff''(0)/2 \epsilon^2 - \bff''(0)/2 \right) \right] \\
	 &= \mathbb{E}\left[ \bff'(0)^2 \epsilon^2 + (\bff'(0) \bff''(0)/2) \epsilon^3 -  (\bff'(0) \bff''(0)/2) \epsilon \right] \\
	 &= \mathbb{E}\left[ \bff'(0)^2 \epsilon^2 \right] \\
	 &= \mathbb{V}[\bff_1(\epsilon)]
\end{align}
where note that $\mathbb{E}[\epsilon^3] = 0$. 
Recall that the optimal control variate can be written
\begin{align}
	c^* &= \mathbb{C}(\bff_1(\epsilon), \bff_2(\epsilon)) / \mathbb{V}[\bff_1(\epsilon)] \\
	&= \mathbb{V}[\bff_1(\epsilon)] / \mathbb{V}[\bff_1(\epsilon)] = 1 \, .
\end{align}

\section{Algorithm Details}
\label{appendix-algorithm}
We summarize an optimization routine using RV-RGE in Algorithm~\ref{alg:reduced-variance}. 
The different variants rely on the different forms of $\bH(\cdot, \cdot)$ and $\text{diag}(\bH)$. 
The \emph{full Hessian} estimator calculates these terms exactly.  
The \emph{diagonal Hessian} estimates the Hessian-vector product with the diagonal of the Hessian.
The \emph{HVP+Local} estimator computes the Hessian-vector product exactly, but estimates the scale approximation mean using other samples.

We also note that there are ways to optimize the additional Hessian-vector product computation.  
Because each Hessian is evaluated at the same $\bmm_\blambda$, we can cache the computation in the forward pass, and only repeat the backwards pass for each sample, as implemented in \cite{autograd}.

\section{Model Definitions}
\subsection{Multi-level Poisson GLM}
\label{sec:appendix-frisk}
Our second test model is a 37-dimensional posterior resulting from a  hierarchical Poisson GLM.
This model measures the relative rates of stop-and-frisk events for different ethnicities and in different precincts \cite{gelman2007analysis}, and has been used as illustrative example of multi-level modeling \cite[Chapter~15, Section~1]{gelman2006data}.

\begin{align*}
	\mu &\sim \mathcal{N}(0, 10^2)  && \text{ mean offset } \\
	\ln \sigma^2_{\alpha}, \ln \sigma^2_{\beta} &\sim \mathcal{N}(0, 10^2) && \text{ group variances } \\
	\alpha_{e} &\sim \mathcal{N}(0, \sigma^2_{\alpha}) && \text{ ethnicity effect }\\
	\beta_{p} &\sim \mathcal{N}(0, \sigma^2_{\beta}) && \text{ precinct effect } \\
	\ln \lambda_{ep} &= \mu + \alpha_{e} + \beta_{p} + \ln N_{ep} && \text{ log rate }\\
	Y_{ep} &\sim \mathcal{P}(\lambda_{ep})  && \text{ stop-and-frisk events }
\end{align*}
where $Y_{ep}$ are the number of stop-and-frisk events within ethnicity group~$e$ and precinct~$p$ over some fixed period of time;~$N_{ep}$ is the total number of arrests of ethnicity group~$e$ in precinct~$p$ over the same period of time;~$\alpha_e$ and~$\beta_p$ are the ethnicity and precinct effects.

\subsection{Bayesian Neural Network}
\label{sec:appendix-bnn}
We implement a 50-unit hidden layer neural network with ReLU activation functions.
We place a normal prior over each weight in the neural network, governed by the same variance (with an inverse Gamma prior).  We also place an inverse Gamma prior over the observation variance
The model can be written as
\begin{align}
	\alpha &\sim \text{Gamma}(1, .1) & \text{ weight prior hyper }\\
	\tau   &\sim \text{Gamma}(1, .1) & \text{ noise prior hyper }\\
	w_{i}  &\sim \mathcal{N}(0, 1/\alpha) &\text{ weights } \\
    y | x, w, \tau &\sim \mathcal{N}( \phi(x, w), 1/\tau ) &\text{ output distribution }
\end{align}
where $w = \{ w \}$ is the set of weights, and $\phi(x, w)$ is a multi-layer perceptron that maps input $x$ to approximate output $y$ as a function of parameters $w$.  We denote the set of parameters as $\theta \triangleq (w, \alpha, \tau)$.  
We approximate the posterior $p(w, \alpha, \tau | \mathcal{D})$, where $\mathcal{D}$ is the training set of $\{ x_n, y_n \}_{n=1}^N$ input-output pairs.

We use a 100-row subsample of the \texttt{wine} dataset from the UCI repository \url{https://archive.ics.uci.edu/ml/datasets/Wine+Quality}. 

\newpage
\section{Variance Reduction}
\label{appendix-variance}
Below are additional variance reduction measurements for the \texttt{frisk} model for different values of $L$, samples drawn per iteration. 

\begin{table}[h!]
  \centering
  \caption{\texttt{frisk} model variance comparison: $L=3$-sample estimators}
  \label{table:variance}
  
\begin{tabular}{l l r r r r r r}
  \toprule
            &              & \multicolumn{2}{c}{$\bg_{\bmm_\blambda}$} & \multicolumn{2}{c}{$\ln \bg_{\bs_\blambda}$} & \multicolumn{2}{c}{$\bg_{\blambda}$} \\
  Iteration & Estimator    & Ave $\mathbb{V}(\cdot)$ & $\mathbb{V}( ||\cdot||)$ & Ave $\mathbb{V}(\cdot)$ & $\mathbb{V}( ||\cdot||)$ & Ave $\mathbb{V}(\cdot)$ & $\mathbb{V}( ||\cdot||)$ \\
  \midrule
  
  \multirow{4}{*}{ early }
    & MC           & 100.000 & 100.000 & 100.000 & 100.000 & 100.000 & 100.000 \\
    & Full Hessian & 1.184 & 1.022 & 0.001 & 0.002 & 0.007 & 0.902 \\
    & Hessian Diag & 35.541 & 25.012 & 0.003 & 0.011 & 0.201 & 22.090 \\
    & HVP + Local  & 1.184 & 1.022 & 0.012 & 0.039 & 0.019 & 0.900 \\
  \midrule

  \multirow{4}{*}{ mid }
    & MC           & 100.000 & 100.000 & 100.000 & 100.000 & 100.000 & 100.000 \\
    & Full Hessian & 0.080 & 0.075 & 0.122 & 0.169 & 0.081 & 0.075 \\
    & Hessian Diag & 39.016 & 22.832 & 6.617 & 8.097 & 38.868 & 22.804 \\
    & HVP + Local  & 0.080 & 0.075 & 31.992 & 46.160 & 0.227 & 0.078 \\
  \midrule

  \multirow{4}{*}{ late }
    & MC           & 100.000 & 100.000 & 100.000 & 100.000 & 100.000 & 100.000 \\
    & Full Hessian & 0.044 & 0.024 & 1.782 & 0.879 & 0.045 & 0.023 \\
    & Hessian Diag & 39.280 & 38.799 & 22.915 & 21.913 & 39.268 & 38.725 \\
    & HVP + Local  & 0.044 & 0.024 & 98.290 & 99.679 & 0.116 & 0.014 \\
  \bottomrule

\end{tabular}

\end{table}

\begin{table}[h!]
  \centering
  \caption{\texttt{frisk} model variance comparison: $L=50$-sample estimators}
  \label{table:variance}
  
\begin{tabular}{l l r r r r r r}
  \toprule
            &              & \multicolumn{2}{c}{$\bg_{\bmm_\blambda}$} & \multicolumn{2}{c}{$\ln \bg_{\bs_\blambda}$} & \multicolumn{2}{c}{$\bg_{\blambda}$} \\
  Iteration & Estimator    & Ave $\mathbb{V}(\cdot)$ & $\mathbb{V}( ||\cdot||)$ & Ave $\mathbb{V}(\cdot)$ & $\mathbb{V}( ||\cdot||)$ & Ave $\mathbb{V}(\cdot)$ & $\mathbb{V}( ||\cdot||)$ \\
  \midrule
  
  \multirow{4}{*}{ early }
    & MC           & 100.000 & 100.000 & 100.000 & 100.000 & 100.000 & 100.000 \\
    & Full Hessian & 1.276 & 1.127 & 0.001 & 0.002 & 0.008 & 1.080 \\
    & Hessian Diag & 35.146 & 24.018 & 0.003 & 0.012 & 0.197 & 23.028 \\
    & HVP + Local  & 1.276 & 1.127 & 0.013 & 0.039 & 0.020 & 1.079 \\
  \midrule

  \multirow{4}{*}{ mid }
    & MC           & 100.000 & 100.000 & 100.000 & 100.000 & 100.000 & 100.000 \\
    & Full Hessian & 0.081 & 0.074 & 0.125 & 0.121 & 0.081 & 0.074 \\
    & Hessian Diag & 37.534 & 21.773 & 7.204 & 7.035 & 37.394 & 21.752 \\
    & HVP + Local  & 0.081 & 0.074 & 31.278 & 32.275 & 0.225 & 0.076 \\
  \midrule

  \multirow{4}{*}{ late }
    & MC           & 100.000 & 100.000 & 100.000 & 100.000 & 100.000 & 100.000 \\
    & Full Hessian & 0.042 & 0.043 & 1.894 & 0.296 & 0.044 & 0.043 \\
    & Hessian Diag & 39.972 & 101.263 & 24.450 & 27.174 & 39.961 & 101.019 \\
    & HVP + Local  & 0.042 & 0.043 & 98.588 & 99.539 & 0.112 & 0.033 \\
  \bottomrule

\end{tabular}

\end{table}

\end{document}